\documentclass[11pt,a4paper]{article}
\usepackage[hyperref]{acl2021}
\usepackage{times}
\usepackage{latexsym}

\usepackage{microtype}

\usepackage{amsmath}
\usepackage{algorithm}
\usepackage{algorithmic}
\usepackage{graphicx}

\newcommand{\KL}{\mathrm{KL}}
\newcommand{\SepaCVAE}{\textsf{SepaCVAE}}
\usepackage{xcolor} 

\definecolor{darkbrown}{rgb}{0.7,0.2,0.1}

\definecolor{orange}{rgb}{1,0.5,0}

\definecolor{darkgreen}{rgb}{0,0.5,0}
\definecolor{grey}{rgb}{0.7,0.7,0.7}

\aclfinalcopy 

\title{Generating Relevant and Coherent Dialogue Responses using Self-separated Conditional Variational AutoEncoders}

\author{Bin Sun$^1$, Shaoxiong Feng$^1$, Yiwei Li$^1$, Jiamou Liu$^2$, Kan Li$^{1*}$\\
  $^1$School of Computer Science and Technology, Beijing Institute of Technology \\
  $^2$School of Computer Science, The University of Auckland \\
  \texttt{\{binsun,shaoxiongfeng,liyiwei,likan\}@bit.edu.cn} \\
  \texttt{jiamou.liu@auckland.ac.nz}}

\date{}

\begin{document}
\maketitle
\begin{abstract}
Conditional Variational AutoEncoder (CVAE) effectively increases the diversity and informativeness of responses in open-ended dialogue generation tasks through enriching the context vector with sampled latent variables. However, due to the inherent \textit{one-to-many} and \textit{many-to-one} phenomena in human dialogues, the sampled latent variables may not correctly reflect the contexts' semantics, leading to irrelevant and incoherent generated responses. To resolve this problem, we propose \textit{Self-separated Conditional Variational AutoEncoder} (abbreviated as \SepaCVAE) that introduces group information to regularize the latent variables, which enhances CVAE by improving the responses' relevance and coherence while maintaining their diversity and informativeness. \SepaCVAE{} actively divides the input data into groups, and then widens the absolute difference between data pairs from distinct groups, while narrowing the relative distance between data pairs in the same group. Empirical results from automatic evaluation and detailed analysis demonstrate that \SepaCVAE{} can significantly boost responses in well-established open-domain dialogue datasets.
\end{abstract}

\section{Introduction}
When conversing with a human user, an open-domain dialogue system is expected to generate human-like responses -- responses that not only  are diverse and informative, but also contain relevant and cohesive information that correctly addresses the context dialogue. Through using sampled latent variables, Conditional Variational AutoEncoders (CVAE) are powerful tools to ensure diversity and informativeness of the generated responses \citep{VaeTextGeneration-Bowman-2016,VHRED-Serban-2017,CVAE(SPhred)-ShenXiaoyu-2017,kgCVAE-ZhaoTiancheng-2017,HVaeMN-ChenHongshen-2018}. Yet, it is challenging for a CVAE-based dialogue generation model to keep the responses relevant and coherent.  The challenge arises as human dialogues inherently exhibit the \textit{one-to-many}  and \textit{many-to-one} phenomena \citep{FilteringData-Csaky-2019}, meaning that the same context could lead to very different responses, and different contexts could lead to the same response, respectively.  As a result, the latent variables sampled by CVAE often fail to capture the correct contextual semantics, as shown in Fig.~\ref{fig:example}, leaving open the possibility that similar contexts producing drastically different latent variables.  This has two particular drawbacks:

\begin{figure}[t]
\centering
\includegraphics[width=0.9\columnwidth]{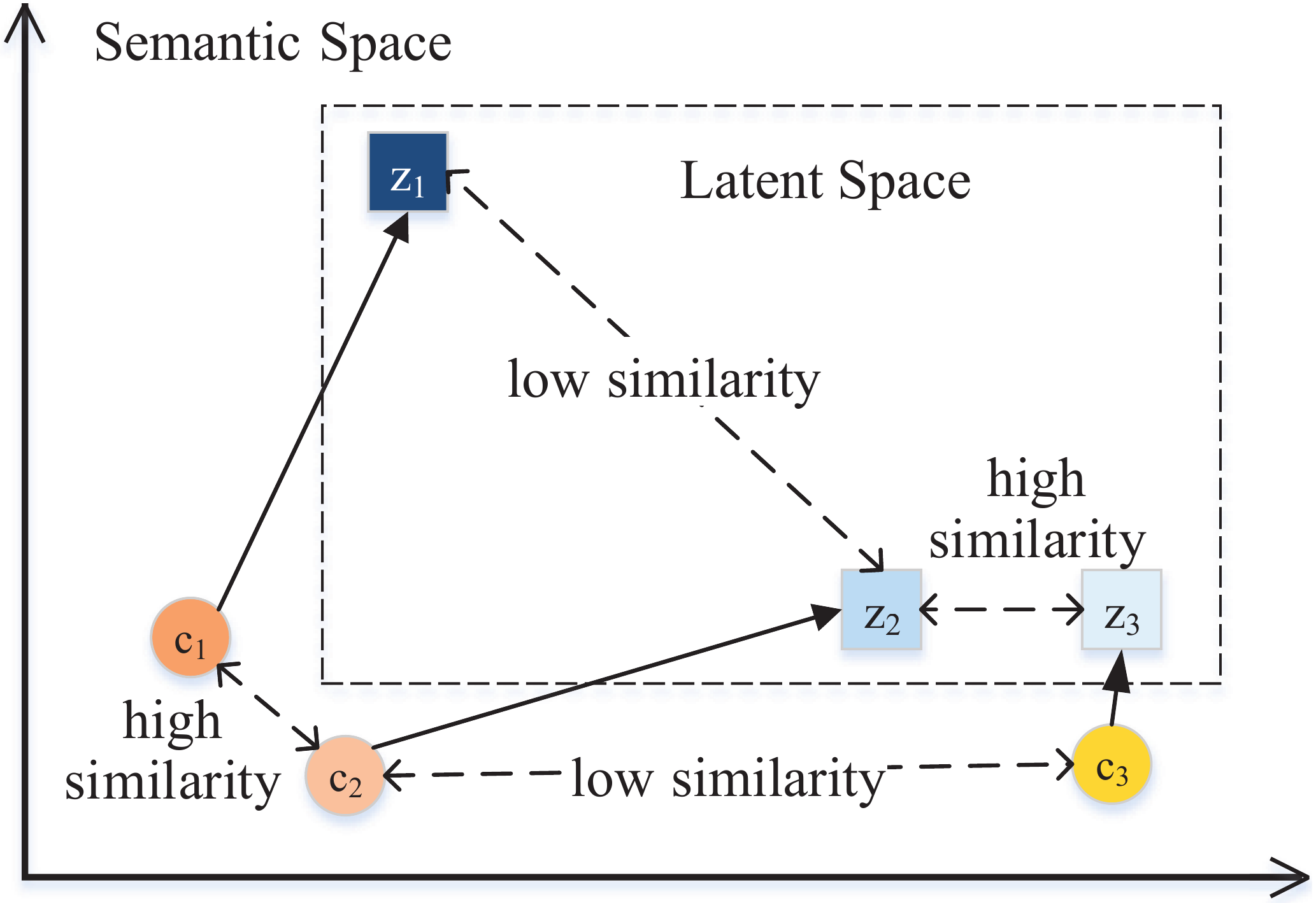}
\caption{In this example, the latent variables $(z_1, z_2,z_3)$ sampled by a general CVAE model don't inherit the semantic relationship of the contexts $(c_1, c_2,c_3)$. Although $c_1$ and $c_2$ have a high similarity, the similarity between $z_1$ and $z_2$ is low. $c_2$ and $c_3$ have a low similarity, but $z_2$ and $z_3$ have a high similarity.}
\label{fig:example}
\label{fig1}
\end{figure}

First, the discrepancy between latent variables could lead to {\em irrelevant and incoherent generated responses}. Different latent variables in a continuous latent space correspond to different responses \citep{VaeTextGeneration-Bowman-2016}. As dissimilar latent variables may be sampled for similar contexts, the generated responses for contexts in the test set could be drastically different from responses to similar contexts in the training set. 
For instance, given a context ``\textit{Everything about this movie is awesome!}'', a standard CVAE may generate response as dissimilar as``\textit{Smartphones of the best games!.}'' and ``\textit{Caves would never say yes, but I’d love to know.}'' \citep{SpaceFusion-GaoXiang-2019}. Thus this approach sacrifices too much relevance and coherence for diversity and informativeness. 

Second, the disparity between contexts and latent variables hurts {\em model generalizability}. Model generalizability is often evaluated using a separate dataset taken from a similar distribution as the training set ({\em e.g.}, a validation or a noisy version of the training set). High generalizability is indicated if the model can transfer favourable abilities from the training set to this second dataset, in the sense  that it produces consistent responses between similar contexts across the two datasets. This suggests that the model has acquired certain semantic relations between sentences from the training set.
However, if the sampled latent variable departs significantly from the contextual semantics, the model may perform quite differently on the second dataset from the training set.

To address these drawbacks, we propose a novel model, namely \textit{Self-Separated Conditional Variational Autoencoder}  (\SepaCVAE). 
\SepaCVAE{} proactively partitions the input data into a number of groups, and then widens the absolute differences between data pairs across different groups while narrowing the relative distance between data pairs within the same group. In this way, \SepaCVAE{} aims to put the contexts that sample similar latent variables into the same groups, thereby regularizing the latent variables.
The design of \SepaCVAE involves three components that are built on top of standard CVAE. First, inspired from image augmentation, we propose a \textit{dialogue augmentation} method to partition data without any prior knowledge. For this, we construct $N$ orthogonal vectors to classify data into $N$ groups, which retain the original semantic relationships of data within a group. We directly enlarge the semantic distance of the data across different groups. Then, we propose a \textit{gradient blocking} algorithm to select the most suitable group for each data according to gains obtained from different groups. Here, the gains are evaluated using reconstruction loss.
Finally, inspired from the {\em contrastive learning} paradigm \citep{GroupwiseCL-CaiHengyi-2020,SimCLR-ChenTing-2020,MoCoV2-ChenXinlei-2020,CLTheory-Jovana-2020}, we propose \textit{relationship enhancement} to increase similarity between the representations of data within the same group, and differentiate the representations of data between different groups. 

\noindent \textbf{Contributions:} Our first contribution is a theoretical analysis on why sampled latent variables fail to reflect the contexts' semantics. The next contribution lies in the proposal of \SepaCVAE{} to overcome issues of irrelevant and incoherent responses caused by standard CVAE. Our third contribution involves a series of experiments. The results show that our \SepaCVAE{} can generate more relevant and coherent responses compared to existing methods.

\section{Related work}
\subsection{Dialogue models}
Open-domain dialogue generation is a challenging task in natural language processing. Early dialogue models \citep{Seq2Seq-ShangLifeng-2015,Seq2Seq-Sordoni-2015} often tend to generate dull responses. To improve the quality of these responses, two pathways have been adopted: one is to introduce external semantic information, such as dialogue history \citep{HRED-QuerySuggest-Sordoni-2015,HRED-Dialogue-Serban-2016}, topic \citep{TopicAware-XingChen-2017}, sentiment \citep{EmotionalTextImage-Huber-2018}, knowledge \citep{FactKnowledge-Ghazvininejad-2018}, persona-style \citep{Persona-LiJiwei-2016}, and other information \citep{MMI-LiJiwei-2016,StyleChatbot-WangDi-2017,DistributionDialog-Baheti-2018,DBLP:conf/emnlp/FengRCSLS20}. The other is through more complex models or frameworks, such as attention mechanisms \citep{Attention-Bahdanau-2015,Attention-ThangLuong-2015},  reinforcement learning (RL) \citep{RLdialoguesys-LiJiwei-2016,RL-Seq2seqCo-Zhang2018,RL-P2BOT-Liu2020}, generative adversarial network (GAN) \citep{SeqGan-YuLantao-2017,AdverREGS-LiJiwei-2017,ConverseGAN(AIM)-ZhangYizhe-2018,PosteriorGan-FengShaoxiong2020}, and variational reasoning \citep{VaeTextGeneration-Bowman-2016,VHRED-Serban-2017,CVAE(SPhred)-ShenXiaoyu-2017,kgCVAE-ZhaoTiancheng-2017,HVaeMN-ChenHongshen-2018}.

CVAE models are conversational models that are based on variational reasoning. Many existing CVAE models have achieved state-of-the-art performance by generating diverse and informative responses. Moreover, as opposed to methods that introduce external semantic information, CVAE models use latent variables to represent such information. Hence they can be applied when external information is not available. Comparing with the models based on RL or GAN, CVAE models are simpler and can be easily trained. In addition, CVAE models can be enhanced by methods that use RL or GAN as generators to further improve their performances.

However, empirical evidences \citep{SpaceFusion-GaoXiang-2019,DialogWAE-GuXiaodong-2019} have indicated that while the  use of latent variables may make the generated responses more diverse and informative, it could also reduce relevance and coherence. To alleviate this apparent issue,  CVAE models have been used in combination with external information such as persona information, dialogue history and dialogue act \citep{CVAE(SPhred)-ShenXiaoyu-2017,VHRED-Serban-2017,kgCVAE-ZhaoTiancheng-2017}.
However, simply borrowing external information is not sufficient to resolve the one-to-many issue, especially when the amount of data is very large. No existing model resolves the core issue of the problem, that is, {\em the latent variable inherits little semantic information from the context sentence}, a consequence of the inherent \textit{one-to-many} and \textit{many-to-one} phenomena of human conversations.
To address this issue, we propose the \SepaCVAE{} model which trains latent variables that inherit contextual semantics.

\subsection{Self-supervised method used in dialogue generation task}
Recently, self-supervised methods such as {\em contrastive learning} -- popularized in computer vision \cite{SimCLR-ChenTing-2020,MoCoV2-ChenXinlei-2020} -- are drawing increasing attention in NLP \citep{SelfSupervisedDL-Jiawei-2019,CL-Clark-2020,GroupwiseCL-CaiHengyi-2020}. 
Generally speaking, the major issue with applying contrastive learning is how positive and negative examples are constructed.
Many existing work explore ways to design reasonable pairs of positive and negative examples to accurately capture the semantic relations of these pairs, so that the obtained representation can be better-used on downstream tasks.

\section{Problem formulation}

The problem with the standard CVAE model lies in that the sampled latent variables may not accurately reflect the contextual semantics due to the apparent \textit{one-to-many} (one context may correspond to many responses) and \textit{many-to-one} (many contexts may also correspond to one response) phenomena. This leads to irrelevant and incoherent responses, and harms model generalizability. Our aim is to adapt sampled latent variables to capture the contextual semantics, so that the effects of these phenomena are neutralized. This will in turn be helpful to generate relevant and coherent responses. With this goal, we focus on {\em single-turn} dialogue datasets where the \textit{one-to-many} situations appear more frequently than multi-turn dialogue datasets.

\subsection{Preconditions}
This section formally analyzes the many-to-one and one-to-many phenomena and we present several important assumptions and contextual information (i.e., preconditions) for the CVAE model.

\noindent \textbf{Notations:} $\theta$ and $\phi$ are parameters of CVAE's recognition network and prior network, respectively; $c$ represents the condition information, $x$ and $r$ represent the generation target, and $z$ represents the latent variable. 

\noindent \textbf{Precondition 1:}
\citet{VaeTextGeneration-Bowman-2016} confirmed that the latent space is continuous; the latent variable $z$ is highly correlated with the target data $x$, meaning that different $z$ will reconstruct different $x$.  

\noindent \textbf{Precondition 2:}
CVAE has a recognition network $q_{\phi}(z|c,x)$ and a prior network $p_{\theta}(z|c)$ to approximate the true posterior distribution $p(z|c,x)$ and prior distribution $p(z|c)$, respectively. These distributions are assumed to follow the Gaussian distribution, \emph{e.g.}, $q_{\phi}(z|c,x)\sim N(\mu, \sigma^2)$.

\noindent \textbf{Precondition 3:}
To efficiently train a CVAE model, the {\em Stochastic Gradient Variational Bayes} (SGVB) framework \citep{ELBO1-Sohn-2015,ELBO2-Yan-2016,StochasticGradientVariationalBayes-kingma-2014} is adopted which aims to maximize the variational lower bound of the conditional log likelihood:
\begin{align}
\label{eq:elbo}
    \nonumber \mathcal{L}(\theta, \phi;c,x) &=-\KL(q_{\phi}(z|c,x)||p_{\theta}(z|c))\\
    &+\mathbf{E}_{q_{\phi}(z|c,x)}\left[\log{p(x|z,c)}\right]
\end{align}
where $\KL$ represents Kullback–Leibler divergence. 
During training, the $\sigma$ of $q(z|x,c)$ will get smaller and smaller, and the $\mu$ of $q(z|x,c)$ will get closer and closer to $z$ that corresponding to $x$, which aims to stabilize the $\mathbf{E}_{q_{\phi}(z|x,c)}\left[\log{p(x|z,c)}\right]$ and make it converge.

\subsection{Demonstrating the existence of the problem}

\begin{figure}[t]
\centering
\includegraphics[width=0.95\columnwidth]{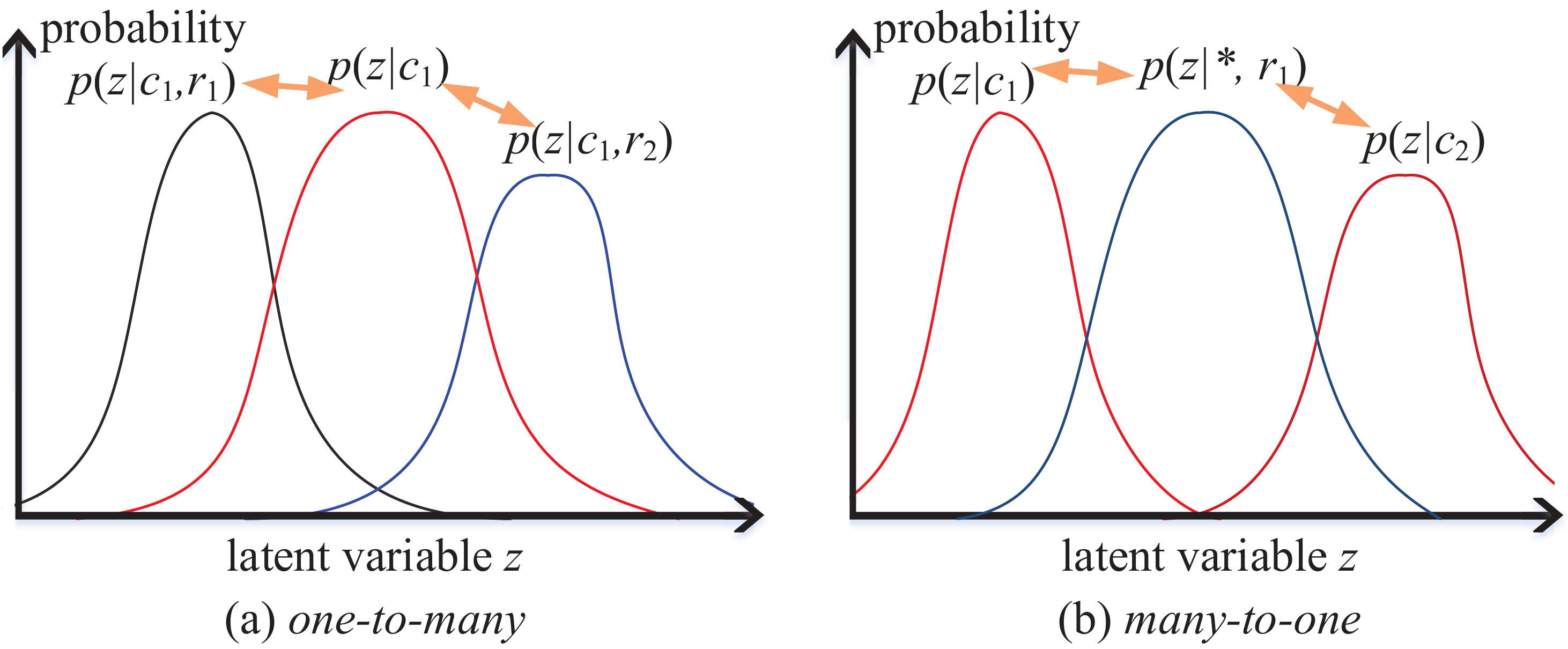}
\caption{The change to the probability distributions of the latent variables of a standard CVAE during training. (a) \textit{one-to-many} phenomenon: Since a context may correspond to two different possible responses $r_1$ and $r_2$, the posterior distributions $p(z|c_1,r_1)$ and $p(z|c_1,r_2)$ are also different. This jeopardizes the requirement of the standard CVAE that these posterior distributions should be similar to the prior distribution $p(z|c_1)$. Therefore, the sampled latent variables from $p(z|c_1)$ may lead to irrelevant and incoherent responses and harm the generalization performance. (b) \textit{many-to-one} phenomenon: Since two different contexts $c_1$ and $c_2$ may have the same response $r_1$, the two prior distributions $p(z|c_1)$ and $p(z|c_2)$ have two corresponding posterior distributions $p(z|c_1,r_1)$ and $p(z|c_2,r_1)$. Since the latent variable $z$ is mainly corresponding to response $r$, $p(z|c_1,r_1)$ and $p(z|c_2,r_1)$ can be assumed as the same, \emph{i.e.}, $p(z|*,r_1)$. Therefore, the prior distributions $p(z|c_1)$ and $p(z|c_2)$ also tend to be the same.}
\label{fig:standard_cvae_training_process}
\end{figure}

We use Fig.~\ref{fig:standard_cvae_training_process} to illustrate the impact of \textit{one-to-many} phenomenon  and \textit{many-to-one} phenomenon on a trained standard CVAE model. 
Consider the situation in Fig.~\ref{fig:standard_cvae_training_process}(a) where the context $c_1$ has two different responses $r_1$ and $r_2$. By \textbf{Precondition 2}, we assume two approximate posterior distributions $p(z|c_1, r_1)\sim N(\mu_1, \sigma_1^2)$, $p(z|c_1, r_2)\sim N(\mu_2, \sigma_2^2)$ and one approximate prior distribution $p(z|c_1)\sim N(\mu, \sigma^2)$. By \textbf{Precondition 3}, during training, $\mu_1$ and $\mu_2$ will get closer  to the latent variables that could be reconstructed to $r_1$ and $r_2$, respectively. By \textbf{Precondition 1}, as $r_1$ is different from $r_2$, $\mu_1$ should also be different from $\mu_2$. Otherwise, the latent variables sampled from $p(z|c_1, r_1)$ and $p(z|c_1, r_2)$ tend to be the same, making these latent variables irrelevant to the responses. This leads to the vanishing latent variable problem \citep{VaeTextGeneration-Bowman-2016}. 
Therefore, $\mu_1$ and $\mu_2$ cannot be the same, and their discrepancy  can be considered stable; only in this way we can ensure one-to-one correspondence between latent variables and responses.
From \textbf{Precondition 3}, it is easy to see that $p(z|c)$ is only affected by $p(z|c,r)$.
Hence, we ignore $\mathbf{E}_{*}\left[\cdot\right]$ in Eq.~(\ref{eq:elbo}) and use $\mathrm{KL}(p(z|c,r)||p(z|c))$ to analyze the trend of $p(z|c)$ during training. Considering Fig.~\ref{fig:standard_cvae_training_process}(a) where $\mathrm{KL}(\cdot)$ of $(c_1,r_1)$ and  $(c_1,r_2)$ equals to $\mathrm{KL}(p(z|c_1, r_1)||p(z|c_1))+\mathrm{KL}(p(z|c_1, r_2)||p(z|c_1))$. We provide details of the computation in \textbf{Appendix~\ref{appendix:problem derivation}}. The formulation can then be simplified as:
$\log\left(\frac{\sigma^2}{\sigma_1\sigma_2}\right) + \frac{\sigma_1^2 + \sigma_2^2+(\mu_1-\mu)^2+(\mu_2-\mu)^2}{2\sigma^2} -1.$

Hence, we can compute $\mu^*$ and $\sigma^*$ that minimizes the above using Lagrange multiplier:
\begin{align}
    &\nonumber \mu^* = (\mu_1+\mu_2)/2\\
    &\nonumber \sigma^* =\sqrt{(\sigma_1^2+\sigma_2^2)/2+(\mu_1-\mu_2)^2/4}.
\end{align}

The derivation above provides insights on the problem caused by the \textit{one-to-many} phenomena in Fig.~\ref{fig:standard_cvae_training_process}(a): 
After training, the prior conditional probability $p(z|c_1) \sim N(\mu^*,\sigma^{*2})$, which will be used in inference. If the difference between $r_1$ and $r_2$ widens, the difference between $\mu_1$ and $\mu_2$ will also widen and $\mu^*$ will become further away from $\mu_1$ and $\mu_2$. During inference, the latent variables sampled from $p(z|c_1)$ have a high probability to differ from those sampled from $p(z|c_1,r_1)$ and $p(z|c_1,r_2)$. These latent variables will introduce irrelevant information and contribute to the generation of irrelevant responses.
In addition, as one response $r_1$ may correspond to different contexts $c_1$ and $c_2$, as shown in Fig.~\ref{fig:standard_cvae_training_process}(b), $p(z|c_1)$ and $p(z|c_2)$ tend to be the same, which contributes to the phenomenon that different context could sample similar latent variables.
In a word, similar contexts could correspond to different latent variables and different contexts could correspond to similar latent variables, which explains  why the latent variables can not accurately reflect the contexts' semantics.

\section{Method}
In this section, we introduce in detail the proposed \SepaCVAE{} model and its three key components, \textit{dialogue augmentation}, \textit{gradient blocking}, and \textit{relationship enhancement}.

\subsection{Self-Separated CVAE}
\begin{figure}[!h]
\centering
\includegraphics[width=0.95\columnwidth]{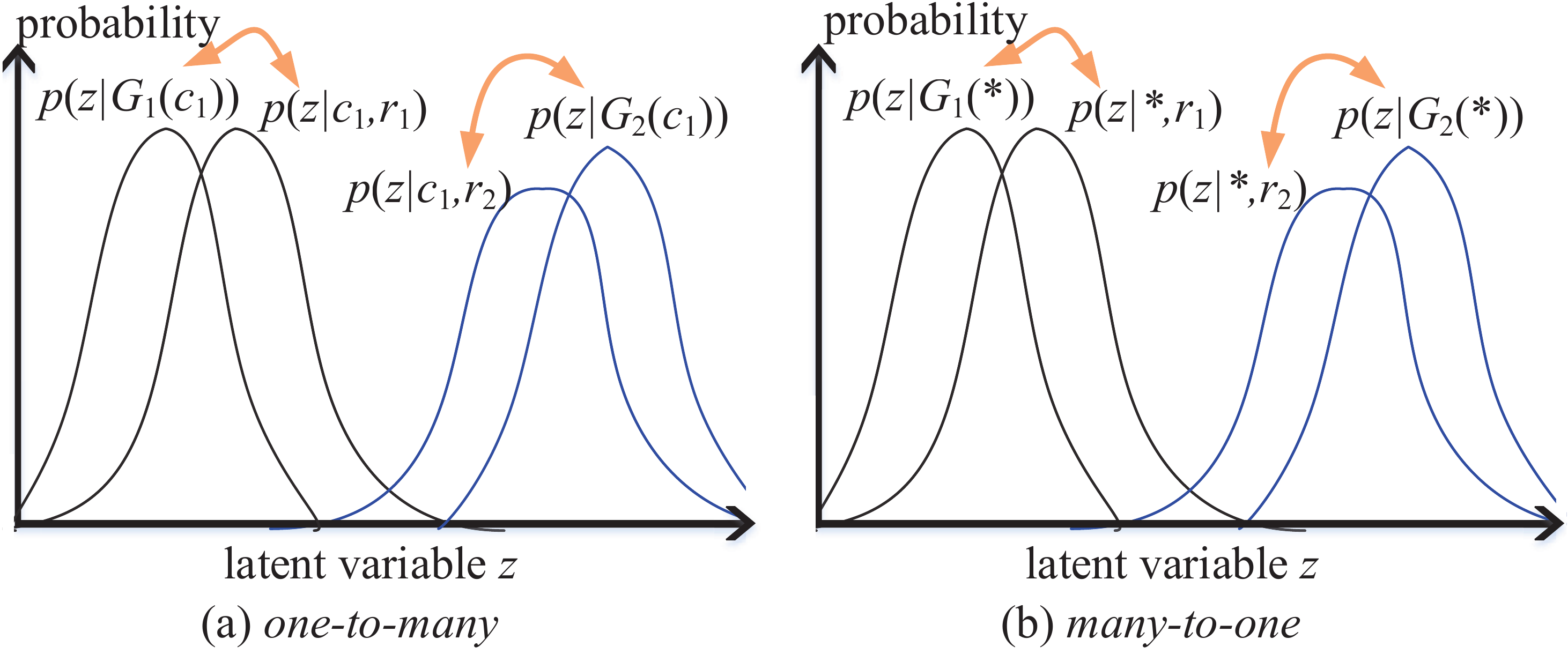}
\caption{Trend of the change of the probability distributions of latent variables of \SepaCVAE{} during training.}
\label{fig:sepacvae_training_process}
\end{figure}
As shown in Fig.~\ref{fig:sepacvae_training_process}, \SepaCVAE{} uses $G(\cdot)$ to separate the contexts into different groups. For the \textit{one-to-many} phenomenon, the contexts in different groups will have different prior distributions $p(z|G_{*}(\cdot))$, which is easily  affected by the different posterior distributions. As for the \textit{many-to-one} phenomenon, \SepaCVAE{} makes the contexts $(c_1, c_2)$ generate latent variables related to the response $r_1$ only when it contains group information $G_1(\cdot)$. The other group would help the contexts to align with the other latent variables.

\subsection{Dialogue augmentation}

In \SepaCVAE{}, we first propose \textit{dialogue augmentation} (see Algorithm~\ref{alg:dialaug}), which designs a group of orthogonal vectors $(y_1, y_2,\ldots , y_N)$ to separate the contexts into different groups. These vectors $(y_1, y_2, \ldots, y_N)$ are called {\em group information}.

\renewcommand{\algorithmicrequire}{\textbf{Input:}}
\renewcommand{\algorithmicensure}{\textbf{Output:}}
\begin{algorithm}[h]
    \caption{Dialogue augmentation}\label{alg:dialaug}
    \begin{algorithmic}[1]
    \REQUIRE
    $C^{ori}_{1\times m}$ : the vector representation of original context sentence after word embedding process;\\
    $N$ : the hyper-parameter;\\
    $m$ : the dimension of word embedding;
    \ENSURE
    $C^{ext}_{N\times m}$ : vector representations of context sentences after augmentation;\\
    $Y^{ext}_{N\times 1}$ : the labels of the augmented contexts;
        \STATE Initialize $C^{ext}_{N\times m}$ and $Y^{ext}_{N\times 1}$;
        \STATE Set $d \leftarrow $ the integer of $m/N$;
        \FOR {$i=1$ to $N$}
        \STATE Initialize augment vector $y_i\leftarrow (0,0,\ldots,0)_{1\times m}$;
        \STATE Set $y_i((i-1)\times d+1:i\times d)\leftarrow (1,1,\ldots,1)_{1\times d}$;
        \STATE $C^{ext}_{N\times m}(i,:)\leftarrow C_{1\times m}^{ori}+y_i$;
        \STATE $Y^{ext}_{N\times 1}(i)\leftarrow i$;
        \ENDFOR
    \RETURN $C^{ext}_{N\times m}$, $Y^{ext}_{N\times 1}$
    \end{algorithmic}
\end{algorithm}

In \SepaCVAE{}, we apply Algorithm~\ref{alg:dialaug} to extend each dialogue pair $(c_i, r_i)$ to $[(c_i+y_1, r_i), (c_i+y_2, r_i), \ldots, (c_i+y_N, r_i)]$ before feeding them to start training. If different contexts $c_i, c_j,\ldots$ have the same $y_i$ added, then these contexts belong to the same group. In this way, 
all contexts will keep a certain relationship within the same group. In this work, the value $N$ is set to 8. Since we use $c+y$ to replace the original $c$, the variational lower bound of \SepaCVAE{} is re-written as:
\begin{align}
    \nonumber \mathcal{L}(\theta, \phi;&r,c,y)=\mathbf{E}_{q_{\phi}(z|r,c+y)}[\log{p(r|z,c+y)}]\\
    &-KL(q_{\phi}(z|r,c+y)||P_{\theta}(z|c+y))
\end{align}

\subsection{Gradient blocking}
Before the gradient back-propagation, we propose \textit{gradient blocking} (see Algorithm~\ref{alg:GradientBlock} in \textbf{Appendix~\ref{appendix:gradient block}} for implementation details) to filter the gradients. Since we extend the dialogue pair $(c, r)$ to $[(c+y_1, r), (c+y_2, r), \ldots, (c+y_N, r)]$, if we optimize the model through all calculated gradients,  $y_1, y_2, \ldots, y_N$ would be regarded as noise. Therefore, We choose the largest variational lower bound that is calculated through the dialogue pair $(c, r)$ with the positive group information $y^+$, which can be represented as \eqref{eq:cvae-loss}:
\begin{equation}
\label{eq:cvae-loss}
    \mathcal{L}(\theta, \phi;r,c,y^+) = \max_{\theta, \phi, y_i\in Y}{\mathcal{L}(\theta, \phi;r,c,y_i)}
\end{equation}
For each $[(c+y_1, r), (c+y_2, r), \ldots, (c+y_N, r)]$, we only pass 
$\mathcal{L}(\cdot,y^+)$ to optimize the model.

\subsection{Relationship enhancement}

Through  \textit{dialogue augmentation} and \textit{gradient blocking}, the positive $y^+$ for each dialogue pair $(c, r)$ is captured. We then propose \textit{relationship enhancement}, which is inspired from \textit{contrastive learning}, to adjust the separated results. Those responses under the same $y^+$ are considered to be in the same group, and thus can be seen as positive samples; similarly, those responses under different $y^+$ are seen as negative samples. From the perspective of contrastive learning, we design a \textit{relationship-enhancement-loss} named $\mathcal{L}_{re}$ to help our model achieve the representation learning:
\begin{align}
 &\mathcal{L}_{re} =\\ \nonumber&-\log\frac{e^{\sum_{j=1}^{Pos}f(x^{'}_i)^T f(x^{'+}_j)}}{e^{\sum_{j=1}^{Pos}f(x^{'}_i)^T f(x^{'+}_j)}+e^{\frac{\sum_{m=1}^{Neg}f(x^{'}_i)^T f(x^{'-}_m)}{N-1}}},
\end{align}
where $x^{'}$ represents the embedded generated response, $f(\cdot)$ represents our model' encoder, $Pos$ means the number of positive samples, and $Neg$ means the number of negative samples.

In addition, we introduce an MLP to predict $y^+$ based on vector representation of the generated response $f(x^{'})$. We therefore define $\mathcal{L}_Y$:
\begin{align}
\mathcal{L}_Y = E_{p_{\psi}(x|z,c+y^+)}\left[\log(p(y^+|x^{'}))\right]
\end{align}

Overall, \SepaCVAE{} is trained by maximizing:
\begin{align}
\mathcal{L}_{all} = \mathcal{L}(\theta,\phi;r,c,y^+) - \alpha * \mathcal{L}_{re} - \mathcal{L}_Y
\end{align}

Quoting the KL annealing trick \citep{VaeTextGeneration-Bowman-2016}, $\alpha$ increases linearly from 0 to 1 in the first 10,000 batches.

\section{Experiments}
\subsection{Dataset}
We use two public dialogue datasets in our experiments, and change them as single-turn dialog data.
The first dataset, named DailyDialog ~\cite{dailydialog2017}, consists of dialogues that resemble human daily communication. The second dataset, named OpenSubtitles ~\cite{opensubtitles2009}, includes a large collection of conversations converted from movie transcripts in English.

\subsection{Data pre-processing}

In this work, we extract single-turn dialogues from two dialogue datasets, DailyDialog and OpenSubtitles. From a multi-turn dialogue $(u_1,u_2,...,u_T)$, we can extract $T-1$ single-turn dialogues $[(u_1,u_2),(u_2,u_3),...,(u_{T-1},u_{T})]$, where $u$ represents an utterance. As discussed above, compared with multi-turn dialogue dataset the single-turn dialogue dataset contains a more serious \textit{one-to-many} problem. Therefore, using the single-turn dialogue dataset for experimentations can highlight the problem of general CVAE model and reflect the effect of our method.

We utilize 300-dimensional GloVe embeddings \cite{GloVe-Pennington-2014} to represent these dialogues in vectors. Since the tokens in GloVe do not cover all tokens in DailyDialog and OpenSubtitles datasets, we extract the token-list of GloVe to filter these datasets.
Table~\ref{tab:data_statistics} lists key statistics of the dataset after processing.
\begin{table}[t]
\centering
\renewcommand\tabcolsep{4.0pt}
\begin{tabular}{lllll}
\hline
    \textbf{dataset name}        & \textbf{vocab}  & \textbf{train} & \textbf{valid} & \textbf{test} \\ \hline
    DailyDialog         & 10,064  & 18,406 & 2,008 & 988 \\
    OpenSubtitles       & 87,840  &  5M    &  100K & 50K   \\
\hline
\end{tabular}
\caption{Statistics for DailyDialog and OpenSubtitles datasets.}
\label{tab:data_statistics}
\end{table}
In addition, we count the \textit{one-to-many} samples of both datasets and found that 408 contexts in DailyDialog and 90,149 contexts in OpenSubtitles have multiple responses. In particular, a context in OpenSubtitles has a maximum of 623 responses, while a context in DailyDialog has a maximum of 29 responses, which shows that the \textit{one-to-many} phenomenon is more prevalent in OpenSubtitles dataset.

\subsection{Automatic evaluation metrics}
\label{sect:dia-met}
We use \textbf{ppl} \citep{ppl}, \textbf{response length} and \textbf{distinct-n} \citep{distinct-16} to evaluate the diversity of generated responses. We also use \textbf{BLEU} \citep{bleu} to evaluate the degree of the word-overlap between generated responses and ground truth.
Moreover, we use \textbf{Embedding Average (Average)}
\citep{embedding-16}) to evaluate the semantic relationship of generated responses and ground-truth responses. Finally, we introduce the \textbf{coherence} \citep{coherence} to assess the coherence between contexts and generated responses.

\subsection{Human evaluation}
We conduct human evaluation to further evaluate our model and baseline models. Following the work of \citet{AdverREGS-LiJiwei-2017,DPGAN-XuJingjing-2018}, we randomly extract 200 samples from the test sets of the two dialogue datasets, respectively. Each sample contains one context and the response generated by different models. Three annotators are invited to rank the generated responses with respect to three aspects: diversity, relevance and fluency. Ties are allowed. Diversity indicates how much the generated response provides specific information, rather than generic and repeated information. Relevance means how likely the generated response is relevant to the context. Fluency specifies how likely the generated response is produced by human.

\subsection{Baseline models}
\label{sect:dia-baselines}

Our baseline models include sequence-to-sequence (Seq2Seq) model, CVAE model, and cluster-CVAE model. They are all implemented based on a 2-layer GRU kgCVAE model \cite{kgCVAE-ZhaoTiancheng-2017}. 
The cluster-CVAE model represents that kgCVAE utilize the cluster results as the knowledge. We employ three cluster methods, \emph{i.e.} K-means(\textbf{K}), Spectral(\textbf{S}), Agglomerative(\textbf{A}).

\subsection{Training details}
\label{sect:dia-para}

\begin{table*}[!t]
\centering
\renewcommand\tabcolsep{3.5pt}
\begin{tabular}{llcccccc}
\hline
  \textbf{mode}             & \textbf{ppl} & \textbf{distinct-1} & \textbf{distinct-2} & \textbf{length} & \textbf{BLEU-1} & \textbf{Average} & \textbf{coherence} \\\hline
  Seq2Seq                 & 42.9$\pm$.18 & 0.033$\pm$.01 & 0.119$\pm$.02 & 9.1$\pm$.22  & 0.386$\pm$.00 & 0.858$\pm$.00 & 0.763$\pm$.00 \\
  CVAE                    & 13.3$\pm$.09 & 0.074$\pm$.00 & 0.407$\pm$.01 & 11.3$\pm$.33 & 0.405$\pm$.01 & 0.853$\pm$.00 & 0.763$\pm$.00 \\
  CVAE+BOW                & 13.0$\pm$.30 & 0.078$\pm$.00 & 0.415$\pm$.01 & 11.4$\pm$.21 & 0.402$\pm$.01 & 0.855$\pm$.00 & 0.762$\pm$.00 \\\hline
  \textbf{K}-CVAE+BOW     & 13.1$\pm$.11 & 0.074$\pm$.00 & 0.406$\pm$.01 & 11.5$\pm$.14 & 0.424$\pm$.00 & \textbf{0.868$\pm$.00} & 0.766$\pm$.00 \\
  \textbf{S}-CVAE+BOW     & 12.9$\pm$.12 & 0.075$\pm$.00 & 0.414$\pm$.01 & 11.5$\pm$.17 & 0.426$\pm$.01 & 0.867$\pm$.00 & 0.765$\pm$.00 \\
  \textbf{A}-CVAE+BOW     & 13.0$\pm$.22 & 0.076$\pm$.00 & 0.418$\pm$.02 & \textbf{11.6$\pm$.11} & 0.418$\pm$.00 & 0.863$\pm$.00 & 0.765$\pm$.00 \\\hline
  SepaCVAE              & \textbf{9.8$\pm$.17}  & \textbf{0.078$\pm$.00} & \textbf{0.504$\pm$.01} & 11.5$\pm$.10 & \textbf{0.461$\pm$.00} & 0.862$\pm$.00 & \textbf{0.767$\pm$.00} \\\hline\hline
  Seq2Seq                & 45.9$\pm$.13 & 0.002$\pm$.00 & 0.010$\pm$.00 & 11.8$\pm$.81 & 0.236$\pm$.04 & 0.465$\pm$.08 & 0.281$\pm$.05 \\
  CVAE+BOW               & 12.2$\pm$.17 & 0.005$\pm$.00 & 0.095$\pm$.00 & \textbf{13.1$\pm$.26} & 0.172$\pm$.02 & 0.285$\pm$.04 & 0.195$\pm$.03 \\\hline
  \textbf{K}-CVAE+BOW    & 12.1$\pm$.20 & 0.006$\pm$.00 & 0.098$\pm$.00 & \textbf{13.1$\pm$.10} & 0.203$\pm$.02 & 0.311$\pm$.06 & 0.200$\pm$.05 \\\hline
  SepaCVAE             & \textbf{2.0$\pm$.06}  & \textbf{0.016$\pm$.00} & \textbf{0.282$\pm$.01} & 12.6$\pm$.11 & \textbf{0.417$\pm$.00} & \textbf{0.836$\pm$.01} & \textbf{0.707$\pm$.01} \\\hline
\end{tabular}
\caption{Metrics results on validation data of DailyDialog (up) and OpenSubtitles (down). The best score in each column is in bold. Note that our BLEU-1 scores are normalized to [0, 1].}
\label{tab:valid_results}
\end{table*}

\begin{table*}[!t]
\centering
\renewcommand\tabcolsep{3.5pt}
\begin{tabular}{lccccccc}
\hline
  \textbf{mode}          & \textbf{distinct-1} & \textbf{distinct-2} & \textbf{length} & \textbf{BLEU-2} & \textbf{BLEU-3} & \textbf{Average} & \textbf{coherence} \\\hline
  Seq2Seq                & 0.054$\pm$.01 & 0.180$\pm$.03 & 9.0$\pm$.32  & 0.300$\pm$.01 & 0.247$\pm$.00 & 0.856$\pm$.00 & 0.756$\pm$.01 \\
  CVAE                   & 0.106$\pm$.00 & 0.499$\pm$.01 & 11.3$\pm$.25 & 0.324$\pm$.01 & 0.272$\pm$.01 & 0.854$\pm$.00 & 0.756$\pm$.00 \\
  CVAE+BOW               & \textbf{0.114$\pm$.00} & \textbf{0.514$\pm$.01} & 11.2$\pm$.13 & 0.326$\pm$.01 & 0.274$\pm$.01 & 0.856$\pm$.00 & 0.755$\pm$.00\\\hline
  \textbf{K}-CVAE+BOW    & 0.108$\pm$.00 & 0.501$\pm$.02 & 11.6$\pm$.16 & 0.342$\pm$.01 & 0.287$\pm$.00 & 0.869$\pm$.00 & 0.759$\pm$.00\\
  \textbf{S}-CVAE+BOW    & 0.110$\pm$.00 & 0.511$\pm$.01 & 11.4$\pm$.19 & 0.339$\pm$.00 & 0.284$\pm$.00 & 0.867$\pm$.00 & 0.758$\pm$.00\\
  \textbf{A}-CVAE+BOW    & 0.111$\pm$.01 & 0.509$\pm$.02 & 11.5$\pm$.16 & 0.331$\pm$.00 & 0.278$\pm$.00 & 0.862$\pm$.00 & 0.757$\pm$.00\\\hline
  SepaCVAE             & 0.082$\pm$.00 & 0.471$\pm$.01 & \textbf{17.9$\pm$.57} & \textbf{0.409$\pm$.01} & \textbf{0.350$\pm$.01} & \textbf{0.877$\pm$.00} & \textbf{0.809$\pm$.00}\\\hline\hline
  Seq2Seq                & 0.003$\pm$.00 & 0.015$\pm$.00 & 11.8$\pm$.82 & 0.193$\pm$.03 & 0.163$\pm$.03 & 0.465$\pm$.08 & 0.281$\pm$.05 \\
  CVAE+BOW               & 0.009$\pm$.00 & 0.131$\pm$.00 & 13.1$\pm$.24 & 0.144$\pm$.02 & 0.123$\pm$.02 & 0.285$\pm$.04 & 0.195$\pm$.03 \\\hline
  \textbf{K}-CVAE+BOW    & 0.010$\pm$.00 & 0.135$\pm$.00 & 13.1$\pm$.10 & 0.169$\pm$.02 & 0.144$\pm$.01 & 0.308$\pm$.06 & 0.198$\pm$.05 \\\hline
  SepaCVAE             & \textbf{0.025$\pm$.00} & \textbf{0.330$\pm$.03} & \textbf{13.5$\pm$.58} & \textbf{0.326$\pm$.01} & \textbf{0.276$\pm$.01} & \textbf{0.807$\pm$.02} & \textbf{0.677$\pm$.01} \\\hline
\end{tabular}
\caption{Mterics results on test data of DailyDialog (up) and OpenSubtitles (down). The best score in each column is in bold. Note that our BLEU-{2,3} scores are normalized to [0, 1].}
\label{tab:test_results}
\end{table*}

For a fair comparison among all models, we utilized 300-dimensional GloVe embeddings as the word embedding matrix.
The numbers of hidden nodes are all set to 300.
The parameter $max\_len$ is set to 25.
We set the batch sizes to 64 and 32 for DailyDialog and OpenSubtitles datasets, respectively.
Adam is utilized for optimization.
The parameter $init\_lr$ is set to 0.001.
We train all models in 50 epochs on a RTX 2080Ti GPU card with Tensorflow, and save the generated responses when the \textbf{ppl} reaching minimum.
Greedy search is used to generate responses for evaluation.

\section{Results and Discussion}
\label{sec:length}

\subsection{Automatic evaluation results}
Table~\ref{tab:valid_results} and Table~\ref{tab:test_results} report the automatic evaluation results of \SepaCVAE{} and baseline models on validation and test data of both two datasets, respectively. For the validation stage, we first select and save the positive group information ($y^+$) for each context, and then generate responses under this $y^+$. For the test data where no ground truth response is available to select the positive group information, we first generate $N$ responses for each context through $N$ group information, and then choose the most possible generated response through calculating the cosine score between the generated responses and context. Both generated responses and contexts are input into \SepaCVAE's encoder to obtain the vector representations. 

\textbf{S}pectral and \textbf{A}gglomerative cluster methods would not work well under the large-scale dataset (\emph{i.e.} OpenSubtitles), and the general CVAE model suffers from the vanishing latent variable problem while training on such dataset.
Therefore, we remove the results of \textbf{S}-CVAE+BOW, \textbf{A}-CVAE+BOW and CVAE on Table~\ref{tab:valid_results} and Table~\ref{tab:test_results}.

As shown in Table~\ref{tab:valid_results} and Table~\ref{tab:test_results}, the results on large-scale dataset (OpenSubtitles) are better than that on small dataset (DailyDialog), that is, the results on OpenSubtitles show an obvious pattern that verifies our hypothesis. On both validation and test data of OpenSubtitles, CVAE and \textbf{K}-CVAE achieve better performance on diversity metric (\textbf{distinct}) but worse performance on relevant metrics (\emph{i.e.} \textbf{BLEU}, \textbf{Average} and \textbf{coherence}) than Seq2Seq model. Moreover, our proposed \SepaCVAE{}  outperforms all baseline models in terms of all metrics with statistical significance.
However, the results obtained on the DailyDialog dataset do not show a clear pattern. For DailyDialog's validation data,  \SepaCVAE{} achieves good performance on diversity but on relevance the results is unimpressive. On the other hand, for test data, \SepaCVAE{} achieves good performance on relevance but generally poor results on diversity. We believe that the reason for this phenomenon is related to the level of prevalence of the \textit{one-to-many} phenomenon in the dataset. For instance, only 66,260 contexts have multiple responses among the 90,149 contexts  on the OpenSubtitles that was added the cluster results. Moreover, one context has a maximum of 296 responses, which amounts to almost half of 623. Since the DailyDialog dataset is very small and contains few samples that we focus on, which cause the not specific tendency on its results. In a word, the evaluation results illustrate the effectiveness of \SepaCVAE{} in terms of improving the relevance and coherence of responses.

\subsection{Human evaluation results}
\begin{table}[!t]
\centering
\small
\begin{tabular}{lccc}
\hline
  \textbf{model}          & \textbf{diversity} & \textbf{relevance} & \textbf{fluency} \\\hline
  Seq2Seq                & 3.64 & 3.12 & \textbf{2.16} \\
  CVAE+BOW               & 3.16 & 3.58 & 3.42 \\
  \textbf{K}-CVAE+BOW    & 3.27 & 3.71 & 3.49 \\
  SepaCVAE               & \textbf{2.11} & \textbf{2.95} & 3.49\\\hline
  Ground-truth           & 1.88 & 1.02 & 1.00 \\\hline\hline
  Seq2Seq                & 3.12 & 3.11 & 3.24 \\
  CVAE+BOW               & 2.69 & 2.98 & 3.05 \\
  \textbf{K}-CVAE+BOW    & 2.59 & 3.53 & 3.72\\
  SepaCVAE               & \textbf{2.57} & \textbf{2.36} & \textbf{2.25}\\\hline
  Ground-truth           & 2.49 & 1.12 & 1.02\\\hline
\end{tabular}
\caption{Human evaluation results on test data of DailyDialog (up) and OpenSubtitles (down). The best score in each column is in bold. Note that ``Ground-truth'' is the true response.}
\label{tab:human_eval}
\end{table}

The results of the human evaluation are shown in Table~\ref{tab:human_eval}. 
To evaluate the consistency of the ranking results assessed by three annotators, we use Pearson's correlation coefficient. 
This coefficient is 0.22 on \textbf{diversity}, 0.63 on \textbf{relevance}, and 0.70 on \textbf{fluency}, with $p < 0.0001$ and below 0.001, which indicates high correlation and agreement. 
Similarly with the automatic evaluation results in Table~\ref{tab:test_results}, this result shows that our \SepaCVAE{} significantly outperforms baselines in term of relevance and diversity. Except the ground-truth responses, our \SepaCVAE{} achieve the best scores of relevance and diversity metrics. The fluency result of \SepaCVAE{} on the DailyDialog dataset is slightly worse than that of baselines, which is mainly due to the length of responses generated by \SepaCVAE{} is almost two times than that of baselines (see Table~\ref{tab:test_results}). When the response lengths are similar on the Opensubtitles dataset, \SepaCVAE{} could also achieve the best fluency score.

\subsection{Effectiveness analysis}

\begin{figure}[!th]
\centering
\includegraphics[width=1.0\columnwidth]{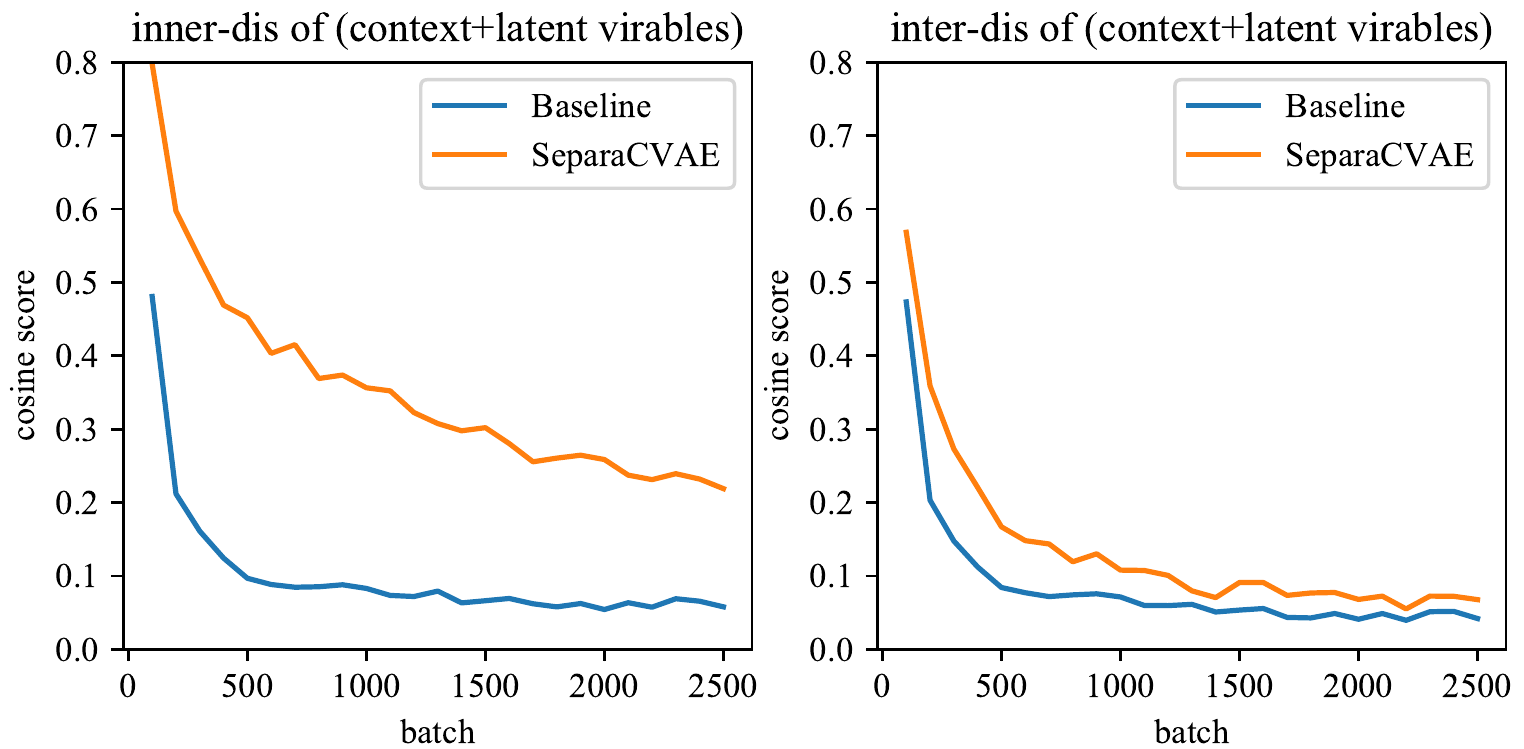}
\caption{The average inner-class distance and the average inter-class distance of the jointly vectors}.
\label{fig:distince_visual}
\end{figure}
\begin{figure}[!th]
\centering
\includegraphics[width=1.0\columnwidth]{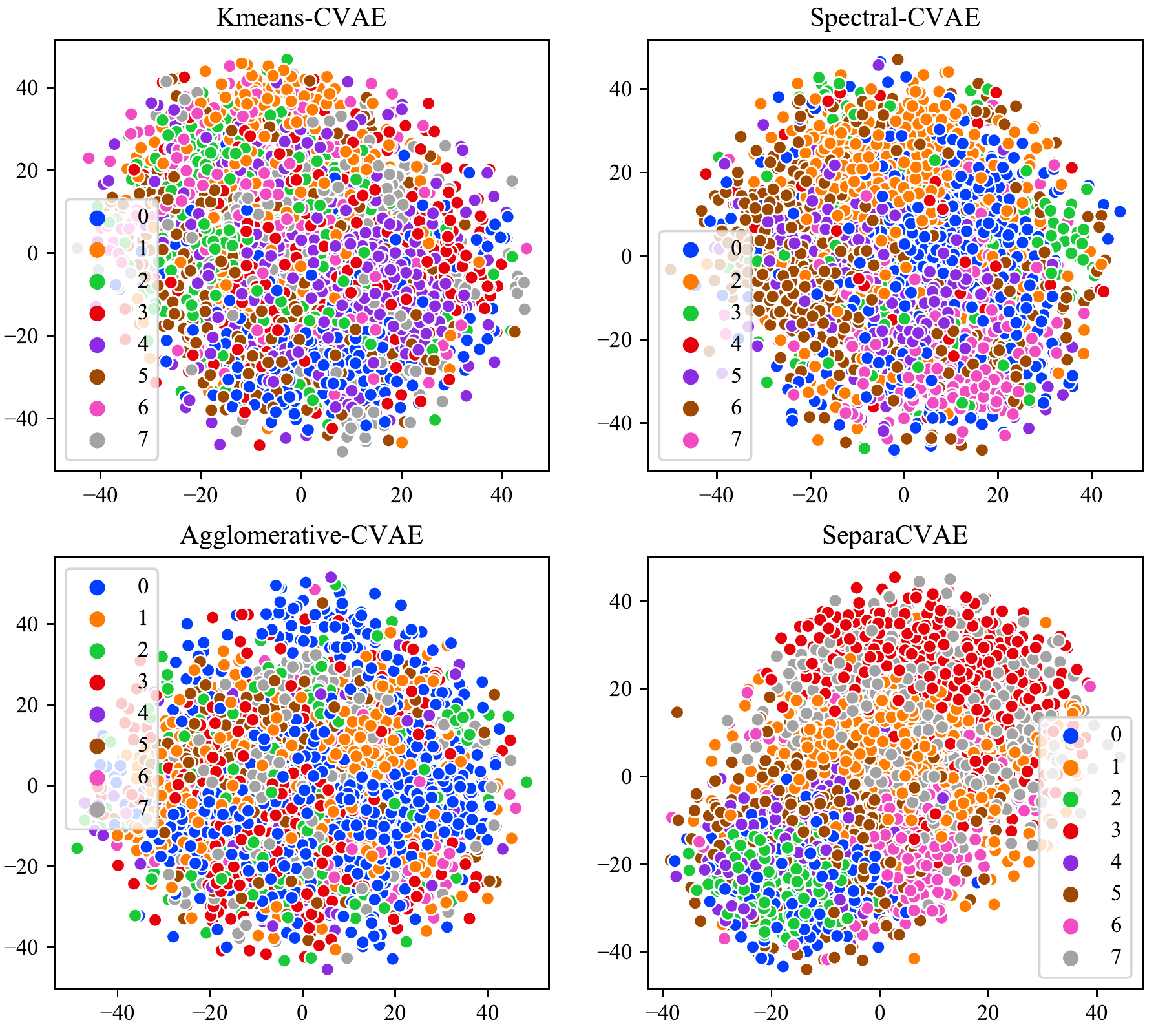} 
\caption{t-SNE visualization of the posterior $z$ for validation responses with 8 group information that obtained though \SepaCVAE{} or cluster methods.}
\label{fig:t-SNE_visual}
\end{figure}

We further analyze the effectiveness of \SepaCVAE{} on regularizing latent variables. For the contexts in the validation data of DailyDialog dataset, we collect their generated responses and the sampled latent variables of both \SepaCVAE{} and baseline models on the first 2,500 batches. Then we calculate the average inner-group distance and the average inter-group distance for each context based on jointly vector representations (concatenating the context vector and the latent variable). All distances are calculated by cosine scores, and the higher the distance, the greater the similarity.

For each context, \SepaCVAE{} outputs a positive group information $y^+$, which is used to distinguish whether other contexts are in the same group.
As for the standard CVAE, we set a threshold of the cosine score to replace the group information. 
In this work, the threshold is set to 0.9.
Finally, we take the average of all contexts' inner-group distance results and inter-group distance results as \textit{inner-dis.} and \textit{inter-dis.} of each batch, which are shown in Fig.~\ref{fig:distince_visual}. \SepaCVAE{} achieves significantly higher \textit{inner-dis.} than baseline (standard CVAE) model, while the \textit{inter-dis.} are similar. Meanwhile, our method also gets the similar average distance of all jointly vectors with the standard CVAE. 

In addition, past studies conjecture that the posterior $z$ sampled from the recognition network should cluster the responses into meaningful groups that correlate with the knowledge. Fig.~\ref{fig:t-SNE_visual} visualizes the posterior $z$ of responses in the validation data of DailyDialog dataset in 2D space using t-SNE \citep{tSNE-Maaten-2008}. We found that the learned latent space of our \SepaCVAE{} is more correlated with the group information. These results demonstrate that \SepaCVAE{} can effectively regularize latent variables.

\subsection{Case study}
\begin{table}[t]
    \centering
    \begin{tabular}{rp{4.4cm}}
    \hline
    & samples in \textbf{training} dataset \\ \hline
    context & would you like to have dinner with me tonight? \\
    true response 1 & oh, that sounds great!\\
    true response 2 & sorry, i have to work overtime.\\\hline\hline
    & sample in \textbf{validation} dataset \\\hline
    similar context & i would always be ready to go shopping with you! should we talk about other basics?	\\\hline
    Seq2Seq & sure. we will go to the movies.\\
    CVAE & i’m not interested in your are not a good thing!	\\
    CVAE+BOW & it will smell and better if whatever, whatever. \\
    SepaCVAE & \textbf{sure, it will be happy,} mary, most music is well.	\\\hline\hline
    & sample in \textbf{test} dataset \\\hline
    similar context & me, too. do you want to go out to celebrate my good news?	\\\hline
    Seq2Seq & yes, i’m going to go to the beach.\\
    CVAE & it really really talking from the street. mom.	\\
    CVAE+BOW & there may live in the rocks, please. \\
    SepaCVAE & \textbf{sure. i go with my parents.} i am so excited about these friends!	\\\hline
    \end{tabular}
    \caption{Generated responses from the baselines and \SepaCVAE.}
    \label{tab:case_study}
\end{table}

We collected the generated responses of contexts in validation and test set, which are similar to the training set, and showed a sample in Table 4. The context in training set has two contradictory responses.
As we analyzed, the standard CVAE and CVAE+BOW generated irrelevant and incoherent response for the similar context in validation and test set. In contrast, our \SepaCVAE{} outputted \textit{sure, it will be happy} and \textit{sure. i go with my parents} are more relevant and coherent than the response generated by baselines, and it also similar with the true response 1 (\textit{oh, that sounds great!}), which means the \SepaCVAE{} is able to handle the \textit{one-to-many} situation.

\section{Conclusion}
In this paper, we theoretically prove that latent variables hardly reflect the semantics of contexts due to the \textit{one-to-many} and \textit{many-to-one} phenomena of dialogues. For the standard CVAE model, these issues lead to irrelevant and incoherent responses during the validation or test stage, and also damaging the generalization performance. To address these problems, we proposed the \SepaCVAE{} model. There are three main technical novelties of \SepaCVAE: dialogue augmentation, gradient blocking, and relationship enhancement, which enable the latent variables to reflect semantic relationships between contexts. As demonstrated in the experimental results, \SepaCVAE{} could get the best performance for large-scale dataset.

\section*{Acknowledgements}
We would like to thank the anonymous reviewers for their constructive comments. This research is supported by Beijing Natural Science Foundation (No. L181010 and 4172054), National Key R\&D Program of China (No. 2016YFB0801100). Kan Li is the corresponding author.

\bibliographystyle{acl_natbib}
\bibliography{anthology,acl2021}

\begin{thebibliography}{48}
\expandafter\ifx\csname natexlab\endcsname\relax\def\natexlab#1{#1}\fi

\bibitem[{Bahdanau et~al.(2015)Bahdanau, Cho, and
  Bengio}]{Attention-Bahdanau-2015}
Dzmitry Bahdanau, Kyunghyun Cho, and Yoshua Bengio. 2015.
\newblock \href {http://arxiv.org/abs/1409.0473} {Neural machine translation by
  jointly learning to align and translate}.
\newblock In \emph{{ICLR}}.

\bibitem[{Baheti et~al.(2018)Baheti, Ritter, Li, and
  Dolan}]{DistributionDialog-Baheti-2018}
Ashutosh Baheti, Alan Ritter, Jiwei Li, and Bill Dolan. 2018.
\newblock \href {https://doi.org/10.18653/v1/d18-1431} {Generating more
  interesting responses in neural conversation models with distributional
  constraints}.
\newblock In \emph{{EMNLP}}, pages 3970--3980. Association for Computational
  Linguistics.

\bibitem[{Bowman et~al.(2016)Bowman, Vilnis, Vinyals, Dai, J{\'{o}}zefowicz,
  and Bengio}]{VaeTextGeneration-Bowman-2016}
Samuel~R. Bowman, Luke Vilnis, Oriol Vinyals, Andrew~M. Dai, Rafal
  J{\'{o}}zefowicz, and Samy Bengio. 2016.
\newblock \href {https://doi.org/10.18653/v1/k16-1002} {Generating sentences
  from a continuous space}.
\newblock In \emph{CoNLL}, pages 10--21. {ACL}.

\bibitem[{Cai et~al.(2020)Cai, Chen, Song, Ding, Bao, Yan, and
  Zhao}]{GroupwiseCL-CaiHengyi-2020}
Hengyi Cai, Hongshen Chen, Yonghao Song, Zhuoye Ding, Yongjun Bao, Weipeng Yan,
  and Xiaofang Zhao. 2020.
\newblock \href {https://doi.org/10.18653/v1/2020.findings-emnlp.70}
  {Group-wise contrastive learning for neural dialogue generation}.
\newblock In \emph{{EMNLP}}, pages 793--802. Association for Computational
  Linguistics.

\bibitem[{Chen et~al.(2018)Chen, Ren, Tang, Zhao, and
  Yin}]{HVaeMN-ChenHongshen-2018}
Hongshen Chen, Zhaochun Ren, Jiliang Tang, Yihong~Eric Zhao, and Dawei Yin.
  2018.
\newblock \href {https://doi.org/10.1145/3178876.3186077} {Hierarchical
  variational memory network for dialogue generation}.
\newblock In \emph{{WWW}}, pages 1653--1662. {ACM}.

\bibitem[{Chen et~al.(2020{\natexlab{a}})Chen, Kornblith, Norouzi, and
  Hinton}]{SimCLR-ChenTing-2020}
Ting Chen, Simon Kornblith, Mohammad Norouzi, and Geoffrey~E. Hinton.
  2020{\natexlab{a}}.
\newblock \href {http://proceedings.mlr.press/v119/chen20j.html} {A simple
  framework for contrastive learning of visual representations}.
\newblock In \emph{{ICML}}, volume 119 of \emph{Proceedings of Machine Learning
  Research}, pages 1597--1607. {PMLR}.

\bibitem[{Chen et~al.(2020{\natexlab{b}})Chen, Fan, Girshick, and
  He}]{MoCoV2-ChenXinlei-2020}
Xinlei Chen, Haoqi Fan, Ross~B. Girshick, and Kaiming He. 2020{\natexlab{b}}.
\newblock \href {https://arxiv.org/abs/2003.04297} {Improved baselines with
  momentum contrastive learning}.
\newblock \emph{CoRR}, abs/2003.04297.

\bibitem[{Clark et~al.(2020)Clark, Luong, Le, and Manning}]{CL-Clark-2020}
Kevin Clark, Minh{-}Thang Luong, Quoc~V. Le, and Christopher~D. Manning. 2020.
\newblock \href {https://openreview.net/forum?id=r1xMH1BtvB} {{ELECTRA:}
  pre-training text encoders as discriminators rather than generators}.
\newblock In \emph{{ICLR}}. OpenReview.net.

\bibitem[{Csaky et~al.(2019)Csaky, Purgai, and
  Recski}]{FilteringData-Csaky-2019}
Richard Csaky, Patrik Purgai, and G{\'{a}}bor Recski. 2019.
\newblock \href {https://doi.org/10.18653/v1/p19-1567} {Improving neural
  conversational models with entropy-based data filtering}.
\newblock In \emph{{ACL} {(1)}}, pages 5650--5669. Association for
  Computational Linguistics.

\bibitem[{Feng et~al.(2020{\natexlab{a}})Feng, Chen, Li, and
  Yin}]{PosteriorGan-FengShaoxiong2020}
Shaoxiong Feng, Hongshen Chen, Kan Li, and Dawei Yin. 2020{\natexlab{a}}.
\newblock \href {https://aaai.org/ojs/index.php/AAAI/article/view/6273}
  {Posterior-gan: Towards informative and coherent response generation with
  posterior generative adversarial network}.
\newblock In \emph{{AAAI}}, pages 7708--7715. {AAAI} Press.

\bibitem[{Feng et~al.(2020{\natexlab{b}})Feng, Ren, Chen, Sun, Li, and
  Sun}]{DBLP:conf/emnlp/FengRCSLS20}
Shaoxiong Feng, Xuancheng Ren, Hongshen Chen, Bin Sun, Kan Li, and Xu~Sun.
  2020{\natexlab{b}}.
\newblock \href {https://doi.org/10.18653/v1/2020.emnlp-main.534} {Regularizing
  dialogue generation by imitating implicit scenarios}.
\newblock In \emph{{EMNLP}}, pages 6592--6604. Association for Computational
  Linguistics.

\bibitem[{Gao et~al.(2019)Gao, Lee, Zhang, Brockett, Galley, Gao, and
  Dolan}]{SpaceFusion-GaoXiang-2019}
Xiang Gao, Sungjin Lee, Yizhe Zhang, Chris Brockett, Michel Galley, Jianfeng
  Gao, and Bill Dolan. 2019.
\newblock \href {https://doi.org/10.18653/v1/n19-1125} {Jointly optimizing
  diversity and relevance in neural response generation}.
\newblock In \emph{{NAACL-HLT} {(1)}}, pages 1229--1238. Association for
  Computational Linguistics.

\bibitem[{Ghazvininejad et~al.(2018)Ghazvininejad, Brockett, Chang, Dolan, Gao,
  Yih, and Galley}]{FactKnowledge-Ghazvininejad-2018}
Marjan Ghazvininejad, Chris Brockett, Ming{-}Wei Chang, Bill Dolan, Jianfeng
  Gao, Wen{-}tau Yih, and Michel Galley. 2018.
\newblock \href
  {https://www.aaai.org/ocs/index.php/AAAI/AAAI18/paper/view/16710} {A
  knowledge-grounded neural conversation model}.
\newblock In \emph{{AAAI}}, pages 5110--5117. {AAAI} Press.

\bibitem[{Gu et~al.(2019)Gu, Cho, Ha, and Kim}]{DialogWAE-GuXiaodong-2019}
Xiaodong Gu, Kyunghyun Cho, Jung{-}Woo Ha, and Sunghun Kim. 2019.
\newblock \href {https://openreview.net/forum?id=BkgBvsC9FQ} {Dialogwae:
  Multimodal response generation with conditional wasserstein auto-encoder}.
\newblock In \emph{{ICLR} (Poster)}. OpenReview.net.

\bibitem[{Huber et~al.(2018)Huber, McDuff, Brockett, Galley, and
  Dolan}]{EmotionalTextImage-Huber-2018}
Bernd Huber, Daniel~J. McDuff, Chris Brockett, Michel Galley, and Bill Dolan.
  2018.
\newblock \href {https://doi.org/10.1145/3173574.3173851} {Emotional dialogue
  generation using image-grounded language models}.
\newblock In \emph{{CHI}}, page 277. {ACM}.

\bibitem[{Kingma and
  Welling(2014)}]{StochasticGradientVariationalBayes-kingma-2014}
Diederik~P. Kingma and Max Welling. 2014.
\newblock \href {http://arxiv.org/abs/1312.6114} {Auto-encoding variational
  bayes}.
\newblock In \emph{{ICLR}}.

\bibitem[{Li et~al.(2016{\natexlab{a}})Li, Galley, Brockett, Gao, and
  Dolan}]{MMI-LiJiwei-2016}
Jiwei Li, Michel Galley, Chris Brockett, Jianfeng Gao, and Bill Dolan.
  2016{\natexlab{a}}.
\newblock \href {https://doi.org/10.18653/v1/n16-1014} {A diversity-promoting
  objective function for neural conversation models}.
\newblock In \emph{{HLT-NAACL}}, pages 110--119. The Association for
  Computational Linguistics.

\bibitem[{Li et~al.(2016{\natexlab{b}})Li, Galley, Brockett, Gao, and
  Dolan}]{distinct-16}
Jiwei Li, Michel Galley, Chris Brockett, Jianfeng Gao, and Bill Dolan.
  2016{\natexlab{b}}.
\newblock \href {https://doi.org/10.18653/v1/n16-1014} {A diversity-promoting
  objective function for neural conversation models}.
\newblock In \emph{{HLT-NAACL}}, pages 110--119. {ACL}.

\bibitem[{Li et~al.(2016{\natexlab{c}})Li, Galley, Brockett, Spithourakis, Gao,
  and Dolan}]{Persona-LiJiwei-2016}
Jiwei Li, Michel Galley, Chris Brockett, Georgios~P. Spithourakis, Jianfeng
  Gao, and William~B. Dolan. 2016{\natexlab{c}}.
\newblock \href {https://doi.org/10.18653/v1/p16-1094} {A persona-based neural
  conversation model}.
\newblock In \emph{{ACL} {(1)}}. {ACL}.

\bibitem[{Li et~al.(2016{\natexlab{d}})Li, Monroe, Ritter, Jurafsky, Galley,
  and Gao}]{RLdialoguesys-LiJiwei-2016}
Jiwei Li, Will Monroe, Alan Ritter, Dan Jurafsky, Michel Galley, and Jianfeng
  Gao. 2016{\natexlab{d}}.
\newblock \href {https://doi.org/10.18653/v1/d16-1127} {Deep reinforcement
  learning for dialogue generation}.
\newblock In \emph{{EMNLP}}, pages 1192--1202. {ACL}.

\bibitem[{Li et~al.(2017{\natexlab{a}})Li, Monroe, Shi, Jean, Ritter, and
  Jurafsky}]{AdverREGS-LiJiwei-2017}
Jiwei Li, Will Monroe, Tianlin Shi, S{\'{e}}bastien Jean, Alan Ritter, and Dan
  Jurafsky. 2017{\natexlab{a}}.
\newblock \href {https://doi.org/10.18653/v1/d17-1230} {Adversarial learning
  for neural dialogue generation}.
\newblock In \emph{{EMNLP}}, pages 2157--2169. {ACL}.

\bibitem[{Li et~al.(2017{\natexlab{b}})Li, Su, Shen, Li, Cao, and
  Niu}]{dailydialog2017}
Yanran Li, Hui Su, Xiaoyu Shen, Wenjie Li, Ziqiang Cao, and Shuzi Niu.
  2017{\natexlab{b}}.
\newblock \href {https://www.aclweb.org/anthology/I17-1099/} {Dailydialog: {A}
  manually labelled multi-turn dialogue dataset}.
\newblock In \emph{{IJCNLP(1)}}, pages 986--995. Asian Federation of Natural
  Language Processing.

\bibitem[{Liu et~al.(2016)Liu, Lowe, Serban, Noseworthy, Charlin, and
  Pineau}]{embedding-16}
Chia{-}Wei Liu, Ryan Lowe, Iulian Serban, Michael Noseworthy, Laurent Charlin,
  and Joelle Pineau. 2016.
\newblock \href {https://doi.org/10.18653/v1/d16-1230} {How {NOT} to evaluate
  your dialogue system: An empirical study of unsupervised evaluation metrics
  for dialogue response generation}.
\newblock In \emph{{EMNLP}}, pages 2122--2132. {ACL}.

\bibitem[{Liu et~al.(2020)Liu, Chen, Chen, Lou, Chen, Zhou, and
  Zhang}]{RL-P2BOT-Liu2020}
Qian Liu, Yihong Chen, Bei Chen, Jian{-}Guang Lou, Zixuan Chen, Bin Zhou, and
  Dongmei Zhang. 2020.
\newblock \href {https://doi.org/10.18653/v1/2020.acl-main.131} {You impress
  me: Dialogue generation via mutual persona perception}.
\newblock In \emph{{ACL}}, pages 1417--1427. Association for Computational
  Linguistics.

\bibitem[{Luong et~al.(2015)Luong, Pham, and
  Manning}]{Attention-ThangLuong-2015}
Thang Luong, Hieu Pham, and Christopher~D. Manning. 2015.
\newblock \href {https://doi.org/10.18653/v1/d15-1166} {Effective approaches to
  attention-based neural machine translation}.
\newblock In \emph{{EMNLP}}, pages 1412--1421. The Association for
  Computational Linguistics.

\bibitem[{van~der Maaten and Hinton(2008)}]{tSNE-Maaten-2008}
Laurens van~der Maaten and Geoffrey Hinton. 2008.
\newblock \href {http://jmlr.org/papers/v9/vandermaaten08a.html} {Visualizing
  data using t-sne}.
\newblock \emph{Journal of Machine Learning Research}, 9(86):2579--2605.

\bibitem[{Mitrovic et~al.(2020)Mitrovic, McWilliams, Walker, Buesing, and
  Blundell}]{CLTheory-Jovana-2020}
Jovana Mitrovic, Brian McWilliams, Jacob Walker, Lars Buesing, and Charles
  Blundell. 2020.
\newblock \href {https://arxiv.org/abs/2010.07922} {Representation learning via
  invariant causal mechanisms}.
\newblock \emph{CoRR}, abs/2010.07922.

\bibitem[{Neubig(2017)}]{ppl}
Graham Neubig. 2017.
\newblock \href {http://arxiv.org/abs/1703.01619} {Neural machine translation
  and sequence-to-sequence models: {A} tutorial}.
\newblock \emph{CoRR}, abs/1703.01619.

\bibitem[{Papineni et~al.(2002)Papineni, Roukos, Ward, and Zhu}]{bleu}
Kishore Papineni, Salim Roukos, Todd Ward, and Wei{-}Jing Zhu. 2002.
\newblock \href {https://doi.org/10.3115/1073083.1073135} {Bleu: a method for
  automatic evaluation of machine translation}.
\newblock In \emph{{ACL}}, pages 311--318. {ACL}.

\bibitem[{Pennington et~al.(2014)Pennington, Socher, and
  Manning}]{GloVe-Pennington-2014}
Jeffrey Pennington, Richard Socher, and Christopher~D. Manning. 2014.
\newblock \href {https://doi.org/10.3115/v1/d14-1162} {Glove: Global vectors
  for word representation}.
\newblock In \emph{{EMNLP}}, pages 1532--1543. {ACL}.

\bibitem[{Serban et~al.(2016)Serban, Sordoni, Bengio, Courville, and
  Pineau}]{HRED-Dialogue-Serban-2016}
Iulian~Vlad Serban, Alessandro Sordoni, Yoshua Bengio, Aaron~C. Courville, and
  Joelle Pineau. 2016.
\newblock \href
  {http://www.aaai.org/ocs/index.php/AAAI/AAAI16/paper/view/11957} {Building
  end-to-end dialogue systems using generative hierarchical neural network
  models}.
\newblock In \emph{{AAAI}}, pages 3776--3784. {AAAI} Press.

\bibitem[{Serban et~al.(2017)Serban, Sordoni, Lowe, Charlin, Pineau, Courville,
  and Bengio}]{VHRED-Serban-2017}
Iulian~Vlad Serban, Alessandro Sordoni, Ryan Lowe, Laurent Charlin, Joelle
  Pineau, Aaron~C. Courville, and Yoshua Bengio. 2017.
\newblock \href {http://aaai.org/ocs/index.php/AAAI/AAAI17/paper/view/14567} {A
  hierarchical latent variable encoder-decoder model for generating dialogues}.
\newblock In \emph{{AAAI}}, pages 3295--3301. {AAAI} Press.

\bibitem[{Shang et~al.(2015)Shang, Lu, and Li}]{Seq2Seq-ShangLifeng-2015}
Lifeng Shang, Zhengdong Lu, and Hang Li. 2015.
\newblock \href {https://doi.org/10.3115/v1/p15-1152} {Neural responding
  machine for short-text conversation}.
\newblock In \emph{{ACL} {(1)}}, pages 1577--1586. {ACL}.

\bibitem[{Shen et~al.(2017)Shen, Su, Li, Li, Niu, Zhao, Aizawa, and
  Long}]{CVAE(SPhred)-ShenXiaoyu-2017}
Xiaoyu Shen, Hui Su, Yanran Li, Wenjie Li, Shuzi Niu, Yang Zhao, Akiko Aizawa,
  and Guoping Long. 2017.
\newblock \href {https://doi.org/10.18653/v1/P17-2080} {A conditional
  variational framework for dialog generation}.
\newblock In \emph{{ACL} {(2)}}, pages 504--509. Association for Computational
  Linguistics.

\bibitem[{Sohn et~al.(2015)Sohn, Lee, and Yan}]{ELBO1-Sohn-2015}
Kihyuk Sohn, Honglak Lee, and Xinchen Yan. 2015.
\newblock \href
  {https://proceedings.neurips.cc/paper/2015/hash/8d55a249e6baa5c06772297520da2051-Abstract.html}
  {Learning structured output representation using deep conditional generative
  models}.
\newblock In \emph{{NIPS}}, pages 3483--3491.

\bibitem[{Sordoni et~al.(2015{\natexlab{a}})Sordoni, Bengio, Vahabi, Lioma,
  Simonsen, and Nie}]{HRED-QuerySuggest-Sordoni-2015}
Alessandro Sordoni, Yoshua Bengio, Hossein Vahabi, Christina Lioma, Jakob~Grue
  Simonsen, and Jian{-}Yun Nie. 2015{\natexlab{a}}.
\newblock \href {https://doi.org/10.1145/2806416.2806493} {A hierarchical
  recurrent encoder-decoder for generative context-aware query suggestion}.
\newblock In \emph{{CIKM}}, pages 553--562. {ACM}.

\bibitem[{Sordoni et~al.(2015{\natexlab{b}})Sordoni, Galley, Auli, Brockett,
  Ji, Mitchell, Nie, Gao, and Dolan}]{Seq2Seq-Sordoni-2015}
Alessandro Sordoni, Michel Galley, Michael Auli, Chris Brockett, Yangfeng Ji,
  Margaret Mitchell, Jian{-}Yun Nie, Jianfeng Gao, and Bill Dolan.
  2015{\natexlab{b}}.
\newblock \href {https://doi.org/10.3115/v1/n15-1020} {A neural network
  approach to context-sensitive generation of conversational responses}.
\newblock In \emph{{HLT-NAACL}}, pages 196--205. ACL.

\bibitem[{Tiedemann(2009)}]{opensubtitles2009}
Jörg Tiedemann. 2009.
\newblock \href {https://doi.org/10.1075/cilt.309.19tie} {\emph{News from
  OPUS—A Collection of Multilingual Parallel Corpora with Tools and
  Interfaces}}.

\bibitem[{Wang et~al.(2017)Wang, Jojic, Brockett, and
  Nyberg}]{StyleChatbot-WangDi-2017}
Di~Wang, Nebojsa Jojic, Chris Brockett, and Eric Nyberg. 2017.
\newblock \href {https://doi.org/10.18653/v1/d17-1228} {Steering output style
  and topic in neural response generation}.
\newblock In \emph{{EMNLP}}, pages 2140--2150. Association for Computational
  Linguistics.

\bibitem[{Wu et~al.(2019)Wu, Wang, and Wang}]{SelfSupervisedDL-Jiawei-2019}
Jiawei Wu, Xin Wang, and William~Yang Wang. 2019.
\newblock \href {https://doi.org/10.18653/v1/p19-1375} {Self-supervised
  dialogue learning}.
\newblock In \emph{{ACL} {(1)}}, pages 3857--3867. Association for
  Computational Linguistics.

\bibitem[{Xing et~al.(2017)Xing, Wu, Wu, Liu, Huang, Zhou, and
  Ma}]{TopicAware-XingChen-2017}
Chen Xing, Wei Wu, Yu~Wu, Jie Liu, Yalou Huang, Ming Zhou, and Wei{-}Ying Ma.
  2017.
\newblock \href {http://aaai.org/ocs/index.php/AAAI/AAAI17/paper/view/14563}
  {Topic aware neural response generation}.
\newblock In \emph{{AAAI}}, pages 3351--3357. {AAAI} Press.

\bibitem[{Xu et~al.(2018{\natexlab{a}})Xu, Ren, Lin, and
  Sun}]{DPGAN-XuJingjing-2018}
Jingjing Xu, Xuancheng Ren, Junyang Lin, and Xu~Sun. 2018{\natexlab{a}}.
\newblock \href {https://doi.org/10.18653/v1/d18-1428} {Diversity-promoting
  {GAN:} {A} cross-entropy based generative adversarial network for diversified
  text generation}.
\newblock In \emph{{EMNLP}}, pages 3940--3949. {ACL}.

\bibitem[{Xu et~al.(2018{\natexlab{b}})Xu, Dusek, Konstas, and
  Rieser}]{coherence}
Xinnuo Xu, Ondrej Dusek, Ioannis Konstas, and Verena Rieser.
  2018{\natexlab{b}}.
\newblock \href {https://doi.org/10.18653/v1/d18-1432} {Better conversations by
  modeling, filtering, and optimizing for coherence and diversity}.
\newblock In \emph{{EMNLP}}, pages 3981--3991. {ACL}.

\bibitem[{Yan et~al.(2016)Yan, Yang, Sohn, and Lee}]{ELBO2-Yan-2016}
Xinchen Yan, Jimei Yang, Kihyuk Sohn, and Honglak Lee. 2016.
\newblock \href {https://doi.org/10.1007/978-3-319-46493-0\_47}
  {Attribute2image: Conditional image generation from visual attributes}.
\newblock In \emph{{ECCV} {(4)}}, volume 9908 of \emph{Lecture Notes in
  Computer Science}, pages 776--791. Springer.

\bibitem[{Yu et~al.(2017)Yu, Zhang, Wang, and Yu}]{SeqGan-YuLantao-2017}
Lantao Yu, Weinan Zhang, Jun Wang, and Yong Yu. 2017.
\newblock \href {http://aaai.org/ocs/index.php/AAAI/AAAI17/paper/view/14344}
  {Seqgan: Sequence generative adversarial nets with policy gradient}.
\newblock In \emph{{AAAI}}, pages 2852--2858. {AAAI} Press.

\bibitem[{Zhang et~al.(2018{\natexlab{a}})Zhang, Lan, Guo, Xu, and
  Cheng}]{RL-Seq2seqCo-Zhang2018}
Hainan Zhang, Yanyan Lan, Jiafeng Guo, Jun Xu, and Xueqi Cheng.
  2018{\natexlab{a}}.
\newblock \href {https://doi.org/10.24963/ijcai.2018/635} {Reinforcing
  coherence for sequence to sequence model in dialogue generation}.
\newblock In \emph{{IJCAI}}, pages 4567--4573. ijcai.org.

\bibitem[{Zhang et~al.(2018{\natexlab{b}})Zhang, Galley, Gao, Gan, Li,
  Brockett, and Dolan}]{ConverseGAN(AIM)-ZhangYizhe-2018}
Yizhe Zhang, Michel Galley, Jianfeng Gao, Zhe Gan, Xiujun Li, Chris Brockett,
  and Bill Dolan. 2018{\natexlab{b}}.
\newblock \href
  {https://proceedings.neurips.cc/paper/2018/hash/23ce1851341ec1fa9e0c259de10bf87c-Abstract.html}
  {Generating informative and diverse conversational responses via adversarial
  information maximization}.
\newblock In \emph{NeurIPS}, pages 1815--1825.

\bibitem[{Zhao et~al.(2017)Zhao, Zhao, and
  Esk{\'{e}}nazi}]{kgCVAE-ZhaoTiancheng-2017}
Tiancheng Zhao, Ran Zhao, and Maxine Esk{\'{e}}nazi. 2017.
\newblock \href {https://doi.org/10.18653/v1/P17-1061} {Learning
  discourse-level diversity for neural dialog models using conditional
  variational autoencoders}.
\newblock In \emph{{ACL} {(1)}}, pages 654--664. {ACL}.

\end{thebibliography}

\newpage
\appendix
\section{The computation of prior probability distribution through KL-divergence on the one-to-many situation}
\label{appendix:problem derivation}
We assume that $p(z|c_1, r_1)\sim N(\mu_1, \sigma_1^2)$, $p(z|c_1, r_2)\sim N(\mu_2, \sigma_2^2)$ and $p(z|c_1)\sim N(\mu, \sigma^2)$. Then, we have:
\begin{align}
\nonumber &KL(p(z|c_1,r_1)||p(z|c_1)) \\
\nonumber &= \int p(z|c_1,r_1)\log\frac{p(z|c_1,r_1)}{p(z|c_1)} dz\\
\nonumber &=\int p(z|c_1,r_1)[\log p(z|c_1,r_1)-\log p(z|c_1)]dz\\
\nonumber &=\int p(z|c_1,r_1)[\log\frac{e^{-\frac{(z-\mu_1)^2}{2\sigma_1^2}}}{\sqrt{2\pi\sigma_1^2}}\\
\nonumber &\ \ \ \ -\log\frac{e^{-\frac{(z-\mu)^2}{2\sigma^2}}}{\sqrt{2\pi\sigma^2}}] dz\\
\nonumber &=\int p(z|c_1,r_1)[-\frac{1}{2} \log 2\pi -\log\sigma_1\\
\nonumber &-\frac{(z-\mu_1)^2}{2\sigma_1^2}+\frac{1}{2}\log 2\pi +\log\sigma+\frac{(z-\mu)^2}{2\sigma^2}] dz\\
\nonumber &=\int p(z|c_1,r_1)[\log\frac{\sigma}{\sigma_1}\\
\nonumber &\ \ \ \ +(\frac{(z-\mu)^2}{2\sigma^2}-\frac{(z-\mu_1)^2}{2\sigma_1^2})]dz\\
\nonumber &=\int p(z|c_1,r_1)\log\frac{\sigma}{\sigma_1}dz \\
\nonumber &+ \int p(z|c_1,r_1)\frac{(z-\mu)^2}{2\sigma^2}dz \\
\nonumber &- \int p(z|c_1,r_1)\frac{(z-\mu_1)^2}{2\sigma_1^2}dz.
\end{align}

Since the $\log\frac{\sigma}{\sigma_1}$ is a constant, and the $\int p(z|c_1,r_1)dz=1$, we have:
\begin{equation}
    \nonumber \int p(z|c_1,r_1)\log\frac{\sigma}{\sigma_1}dz=\log\frac{\sigma}{\sigma_1}.
\end{equation}

Since $p(z|c_1,r_1)=\frac{1}{\sqrt{2\pi}\sigma_1}e^{-\frac{(z-\mu_1)^2}{2\sigma^2}}$, the $\int p(z|c_1,r_1)\frac{(z-\mu_1)^2}{2\sigma_1^2}dz$ can be calculated as follow:
\begin{align}
    \nonumber &\int p(z|c_1,r_1)\frac{(z-\mu_1)^2}{2\sigma_1^2}dz\\
    \nonumber &=\int \frac{1}{\sqrt{2\pi}\sigma_1}e^{-\frac{(z-\mu_1)^2}{2\sigma^2}}\frac{(z-\mu_1)^2}{2\sigma_1^2}dz\\
    \nonumber &=\int \frac{1}{\sqrt{2\pi}\sigma_1}e^{-\frac{(z-\mu_1)^2}{2\sigma^2}}\frac{(z-\mu_1)^2}{2\sigma_1^2}\sqrt{2}\sigma_1d\frac{z-\mu_1}{\sqrt{2}\sigma_1}\\
    \nonumber &=\int \frac{1}{\sqrt{\pi}}e^{-\frac{(z-\mu_1)^2}{2\sigma^2}}\frac{(z-\mu_1)^2}{2\sigma_1^2}d\frac{z-\mu_1}{\sqrt{2}\sigma_1}.
\end{align}

Let the $x$=$\frac{z-\mu_1}{\sqrt{2}\sigma_1}$, we have:
\begin{align}
    \nonumber &\int p(z|c_1,r_1)\frac{(z-\mu_1)^2}{2\sigma_1^2}dz\\
    \nonumber &=\frac{1}{\sqrt{\pi}} \int e^{-x^2}x^2dx\\
    \nonumber &=-\frac{1}{2\sqrt{\pi}} \int xde^{-x^2}\\
    \nonumber &=-\frac{1}{2\sqrt{\pi}} (xe^{-x^2}|_{-\infty}^{+\infty}-\int e^{-x^2}dx).
\end{align}

According to the L'Hospital's rule, the $\lim_{x\to-\infty}{xe^{-x^2}}$=$\lim_{x\to+\infty}{xe^{-x^2}}=0$.

To calculate the $\int e^{-x^2}dx$, we first compute the $(\int_0^{+\infty} e^{-x^2}dx)^2$, so we have:
\begin{align}
    \nonumber (\int_0^{+\infty} e^{-x^2}dx)^2 &= \int_0^{+\infty} e^{-x^2}dx \\
    \nonumber &\ \ \ \ \cdot \int_0^{+\infty} e^{-y^2}dy\\
    \nonumber &=\int_0^{+\infty} \int_0^{+\infty} e^{-x^2-y^2}dxdy.
\end{align}

Let $x=r\sin{\theta}$ and $y=r\cos{\theta}$, we have:
\begin{align}
    \nonumber &\int_0^{+\infty} \int_0^{+\infty} e^{-x^2-y^2}dxdy \\
    \nonumber &=\int_0^{\frac{\pi}{2}} \int_0^{+\infty} e^{-r^2}rdrd\theta \\
    \nonumber&= \frac{\pi}{2} \int_0^{+\infty} e^{-r^2}rdr= \frac{\pi}{4}.
\end{align}

Therefore, the $\int_0^{+\infty} e^{-x^2}dx$ = $\frac{\sqrt{\pi}}{2}$. According to the symmetry, the $\int_{-\infty}^{+\infty} e^{-x^2}dx$=$\sqrt{\pi}$. and the $\int p(z|c_1,r_1)\frac{(z-\mu_1)^2}{2\sigma_1^2}dz=\frac{1}{2}$.

For the $\int p(z|c_1,r_1)\frac{(z-\mu)^2}{2\sigma^2}dz$, we have:
\begin{align}
\nonumber &\int p(z|c_1,r_1)\frac{(z-\mu)^2}{2\sigma^2}dz \\
\nonumber &=\int p(z|c_1,r_1)\frac{(z-\mu_1+\mu_1-\mu)^2}{2\sigma^2}dz \\
\nonumber &=\frac{1}{2\sigma^2} [\int (z-\mu_1)^2 p(z|c_1,r_1)dz\\
\nonumber &+\int (\mu_1-\mu)^2 p(z|c_1,r_1) dz \\
\nonumber &+\int (z-\mu_1)(\mu_1-\mu_2)p(z|c_1,r_1)dz] \\
\nonumber &=\frac{2\sigma_1^2\int \frac{(z-\mu_1)^2}{2\sigma_1^2}p(z|c_1,r_1)dz + (\mu_1-\mu)^2}{2\sigma^2} \\
\nonumber &=\frac{\sigma_1^2 + (\mu_1-\mu)^2}{2\sigma^2}.
\end{align}

Therefore, we have:
\begin{align}
    \nonumber &KL(p(z|c_1,r_1)||p(z|c_1)) \\
    \nonumber &= \log\frac{\sigma}{\sigma_1}+\frac{\sigma_1^2 + (\mu_1-\mu)^2}{2\sigma^2}-\frac{1}{2}.
\end{align}

In the same way, the $KL(p(z|c_1,r_2)||p(z|c_1))$ equals $\log\frac{\sigma}{\sigma_2}+\frac{\sigma_2^2 + (\mu_2-\mu)^2}{2\sigma^2}-\frac{1}{2}$. And then, we can know:
\begin{align}
    \nonumber &KL(p(z|c_1, r_1)||p(z|c_1))\\
    \nonumber &\ \ \ \ +KL(p(z|c_1, r_2)||p(z|c_1))\\
    \nonumber &=\log(\frac{\sigma^2}{\sigma_1\sigma_2}) \\
    \nonumber &+ \frac{\sigma_1^2 + \sigma_2^2+(\mu_1-\mu)^2+(\mu_2-\mu)^2}{2\sigma^2} -1.
\end{align} 

Since the Latent Vanish problem is not expected by the VAE and CVAE methods, the $p(z|c_1,r_1)$ should be different from $p(z|c_1,r_2)$, which means the $N(\mu_1,\sigma_1)$ is different from the $N(\mu_2,\sigma_2)$.

After that, we use the $\phi(\mu,\sigma)$ represent the $KL(p(z|c_1, r_1)||p(z|c_1))+KL(p(z|c_1, r_2)||p(z|c_2))$, then we have:
\begin{align}
    \nonumber &\phi(\mu,\sigma)=\log(\frac{\sigma^2}{\sigma_1\sigma_2})\\
    \nonumber &+\frac{\sigma_1^2 + \sigma_2^2+(\mu_1-\mu)^2+(\mu_2-\mu)^2}{2\sigma^2} -1.
\end{align}

According to the Lagrange Multiplier Method, we can calculate the conditional extremum and the extreme point ($\mu^*$,$\sigma^*$) of $\phi(\mu,\sigma)$.

To obtain the $\mu^*$, we have to calculate the $\frac{\partial\phi(\mu,\sigma)}{\partial\mu}$:
\begin{align}
    \nonumber \frac{\partial\phi(\mu,\sigma)}{\partial\mu} &= \frac{\partial\frac{(\mu_1-\mu)^2+(\mu_2-\mu)^2}{2\sigma^2}}{\partial\mu}\\
    \nonumber &= \frac{2\mu-\mu_1-\mu_2}{\sigma^2}.
\end{align}

Let the $\frac{\partial\phi(\mu,\sigma)}{\partial\mu}$ equals 0, we have the $\mu^*$=$\frac{\mu_1+\mu_2}{2}$. In the same way, to obtain the $\sigma^*$, we have:
\begin{align}
    \nonumber &\frac{\partial\phi(\mu,\sigma)}{\partial\sigma} = \frac{\partial\log(\frac{\sigma^2}{\sigma_1\sigma_2})}{\partial\sigma} \\
    \nonumber &+ [\sigma_1^2+\sigma_2^2+(\mu_1-\mu)^2+(\mu_2-\mu)^2]\frac{\partial\frac{1}{2\sigma^2}}{\partial\sigma} \\
    \nonumber &=\frac{2}{\sigma}- \frac{\sigma_1^2+\sigma_2^2+(\mu_1-\mu)^2+(\mu_2-\mu)^2}{\sigma^3}\\
    \nonumber &= \frac{2\sigma^2-[\sigma_1^2+\sigma_2^2+(\mu_1-\mu)^2+(\mu_2-\mu)^2]}{\sigma^3},
\end{align}
where $a$ means the base of the logarithmic formula.

Let the $\frac{\partial\phi(\mu,\sigma)}{\partial\sigma}=0$, since the $\sigma^3$ can not be 0, we have:
\begin{equation}
    \nonumber 2\sigma^2-[\sigma_1^2+\sigma_2^2+(\mu_1-\mu)^2+(\mu_2-\mu)^2] = 0.
\end{equation}

Therefore, the $\sigma^*$ is:
\begin{equation}
   \nonumber \sigma^*=\sqrt{\frac{\sigma_1^2+\sigma_2^2+(\mu_1-\mu)^2+(\mu_2-\mu)^2}{2}}.
\end{equation}

Replace the $\mu$ with the $\mu^*$, we have:
\begin{equation}
   \nonumber \sigma^*=\sqrt{\frac{\sigma_1^2+\sigma_2^2+\frac{(\mu_1-\mu_2)^2}{2}}{2}}.
\end{equation}

We use a constant $C$ to replace $\frac{(\mu_1-\mu_2)^2}{4}$, the $\sigma^*$ equals $\sqrt{\frac{\sigma_1^2+\sigma_2^2}{2}+C}$.

The $\mu^*$=$\frac{\mu_1+\mu_2}{2}$ means the latent variables sampled from this prior probability distribution easily tend to be different from the latent variables sampled form the posterior probability distributions. Since the latent variables are highly correlated with the generated responses, the responses generated through prior probability distribution would be different from that generated from posterior probability distributions. If the difference between $\mu_1$ and $\mu_2$ is very large, the $\sigma^*$ would be large too, thus resulting in high probability of more irrelevant latent variables.

\section{The implementation of \textit{gradient blocking}}
\label{appendix:gradient block}
\begin{algorithm}[!htb]
    \caption{Gradient blocking}\label{alg:GradientBlock}
    \begin{algorithmic}[1]
    \REQUIRE
    $Loss$ : loss-results of extended dialogue data in one batch;\\
    $N$ : the number of group information;\\
    $BatchSize$ : the number of data contained on one Batch;
    \ENSURE
    $Loss\_Mask$ : the mask tensor with [0,1] elements;
        \STATE $Loss\leftarrow $  tf.reshape($Loss$, [$BatchSize$, $N$])
        \STATE $ministLossPOSs\leftarrow $ tf.argmin($Loss$, 1) \# find the posision of the minist loss;
        \STATE $ones\leftarrow $  OnesTensor(1, dtype=tf.float32)
        \STATE $zeros\leftarrow $  ZerosVector(1, dtype=tf.float32)
        \STATE $Loss\_Mask\leftarrow $ tf.cond(\\tf.equal($ministLossPOSs[0]$,\\tf.constant([0])[0],\\ lambda:$ones$, lambda:$zeros$)
        \FOR {$i=1$ to $BatchSize$}
        \FOR {$j=1$ to $N$}
        \IF {$i=1$ and $j = 1$}
        \STATE continue
        \ELSE
        \STATE $Loss\_Mask\leftarrow $  tf.concat([\\$Loss\_Mask$, tf.cond(\\tf.equal($ministLossPOSs[i]$,\\tf.constant([$j$]))[0], lambda:$ones$, lambda:$zeros$)],0)
        \ENDIF
        \ENDFOR
        \ENDFOR
        \STATE $Pass\_Loss \leftarrow Loss$*$Loss\_Mask$
    \RETURN $Pass\_Loss$
    \end{algorithmic}
\end{algorithm}

We present the implementation of \textit{gradient blocking} method in Algorithm~\ref{alg:GradientBlock}.
In Algorithm~\ref{alg:GradientBlock}, we build a mask tensor $Loss\_Mask$ to filter the loss results form each batch data, which can same obstruct the gradient backpropagation. Since we used gradient descent to optimize the neural model, the smallest loss result equals the largest variational lower bound. The elements in $Loss\_Mask$ are 0 or 1, so  $Loss* Loss\_Mask$ can be considered as the selection of the existing $Loss$.

\end{document}